\documentclass[11pt,a4paper]{article}
\usepackage[hyperref]{acl2020}
\usepackage{times}
\usepackage{latexsym}
\usepackage{graphicx}
\usepackage{amsmath}
\usepackage{amssymb}
\usepackage{bbm}
\usepackage{color}
\usepackage{subcaption}
\usepackage{xspace}
\usepackage{url}
\usepackage{booktabs}

\usepackage{microtype}

\aclfinalcopy %
\title{Balancing Objectives in Counseling Conversations: \\Advancing Forwards or Looking Backwards}

\author{Justine Zhang \\
  Cornell University \\
  \texttt{jz727@cornell.edu} \\\And
  Cristian Danescu-Niculescu-Mizil \\
  Cornell University \\
  \texttt{cristian@cs.cornell.edu} \\}

\date{}

\begin{document}

\definecolor{marsala}{RGB}{150, 79, 76}
\definecolor{navy}{RGB}{0, 0, 128}
\definecolor{fgreen}{RGB}{34, 139, 34}
\definecolor{turquoise}{RGB}{69, 181, 170}
\definecolor{lightblue}{RGB}{65,105,225}

\newcommand{\cut}[1]{}

\newcommand{\todo}[1]{\xspace\textcolor{red}{[TODO: #1]}}
\newcommand{\cd}[1]{{\textcolor{turquoise}{[#1]}}}
\newcommand{\jz}[1]{{\textcolor{marsala}{[#1]}}}
\newcommand{\topic}[1]{{\textcolor{blue}{\noindent\bfseries #1.}}}
\newcommand{\sketch}[1]{{\textcolor{lightblue}{#1}}}
\newcommand{\edit}[1]{\textcolor{fgreen}{#1}}

\newcommand{\xhdr}[1]{{\noindent\bfseries #1.}}
\newcommand{\xhdrit}[1]{{\noindent\textit{#1.}}}
\newcommand{\overbar}[1]{\mkern 1.5mu\overline{\mkern-1.5mu#1\mkern-1.5mu}\mkern 1.5mu}

\newcommand{\egword}[1]{{\textit{#1}}\xspace}

\newcommand{\inter}[1]{\ensuremath{\mathcal{#1}}\xspace}

\newcommand{\fwdmessage}{\ensuremath{c_1}\xspace}
\newcommand{\bkmessage}{\ensuremath{c_2}\xspace}
\newcommand{\fwdtxtmessage}{\ensuremath{t_1}\xspace}
\newcommand{\bktxtmessage}{\ensuremath{t_2}\xspace}

\newcommand{\orientation}[1]{\ensuremath{\Omega_{#1}}\xspace}
\newcommand{\maxorientation}[1]{\ensuremath{\Omega_{#1}^{\max}}\xspace}
\newcommand{\minorientation}[1]{\ensuremath{\Omega_{#1}^{\min}}\xspace}

\newcommand{\word}{\ensuremath{w}\xspace}
\newcommand{\fwdword}{\ensuremath{w_1}\xspace}
\newcommand{\bkword}{\ensuremath{w_2}\xspace}
\newcommand{\leftrepr}{\ensuremath{\overleftarrow{w}}\xspace}
\newcommand{\rightrepr}{\ensuremath{\overrightarrow{w}}\xspace}
\newcommand{\wordor}{\orientation{\word}\xspace}

\newcommand{\leftattn}[1]{\ensuremath{\overleftarrow{\sigma_{#1}}}\xspace}
\newcommand{\rightattn}[1]{\ensuremath{\overrightarrow{\sigma_{#1}}}\xspace}
\newcommand{\latentspace}{\ensuremath{\mathbb{T}}\xspace}

\newcommand{\ccmessage}{\ensuremath{c_i}\xspace}
\newcommand{\txmessage}{\ensuremath{t_i}\xspace}
\newcommand{\ccmessages}[1]{\ensuremath{C_{#1}}\xspace}
\newcommand{\leftmessage}{\ensuremath{\overleftarrow{t_i}}\xspace}
\newcommand{\rightmessage}{\ensuremath{\overrightarrow{t_i}}\xspace}
\newcommand{\leftmessages}[1]{\ensuremath{\overleftarrow{T_{#1}}}\xspace}
\newcommand{\rightmessages}[1]{\ensuremath{\overrightarrow{T_{#1}}}\xspace}
\newcommand{\tripleset}{\ensuremath{\{(\leftmessage,\ccmessage,\rightmessage)\}}\xspace}

\newcommand{\txmtx}{\ensuremath{\mathcal{X}}\xspace}
\newcommand{\ccmtx}{\ensuremath{\mathcal{W}}\xspace}

\newcommand{\txrepr}{\ensuremath{U}\xspace}
\newcommand{\lefttxrepr}{\ensuremath{\overleftarrow{U}}\xspace}
\newcommand{\righttxrepr}{\ensuremath{\overrightarrow{U}}\xspace}
\newcommand{\lefttxsub}[1]{\ensuremath{\overleftarrow{U_{#1}}}\xspace}
\newcommand{\righttxsub}[1]{\ensuremath{\overrightarrow{U_{#1}}}\xspace}
\newcommand{\txvect}{\ensuremath{u_i}\xspace}
\newcommand{\leftccrepr}{\ensuremath{\overleftarrow{W}}\xspace}
\newcommand{\rightccrepr}{\ensuremath{\overrightarrow{W}}\xspace}

\newcommand{\bkcentre}[1]{\ensuremath{\overleftarrow{u_{#1}}}\xspace}
\newcommand{\fwdcentre}[1]{\ensuremath{\overrightarrow{u_{#1}}}\xspace}
\newcommand{\wordrepr}[2]{\ensuremath{\text{w}^{#1}_{#2}}\xspace}
\newcommand{\wordreprs}[1]{\ensuremath{\text{w}_{#1}}\xspace}

\newcommand{\intercc}{\inter{C}}
\newcommand{\intertx}{\inter{T}}

\maketitle
\begin{abstract}

Throughout a conversation, participants make choices that can orient the flow of the interaction. 
Such choices are particularly salient in the consequential domain of crisis counseling, 
where a difficulty for counselors is balancing between two 
key objectives:
advancing the conversation towards a resolution, 
and empathetically addressing the crisis situation.

In this work, we develop an unsupervised methodology to quantify how 
counselors manage this balance. 
Our main intuition is that if an utterance can only receive a narrow range of 
appropriate
 replies, then its likely aim is to advance the conversation forwards,
  towards a target within that range.
Likewise, an utterance that can only appropriately follow a narrow range of possible utterances is likely aimed backwards at addressing a specific situation within that range.
By applying this intuition, we can map each utterance to a continuous \emph{orientation} axis that captures the degree to which it is intended to 
direct the flow of the conversation forwards or backwards.

This unsupervised method allows us to characterize counselor behaviors in a large dataset of crisis counseling conversations, where we show that known counseling strategies intuitively align with this axis. 
We also illustrate how our measure can be indicative of a conversation's progress, as well as its effectiveness.

\end{abstract}

\section{Introduction}
\label{sec:intro}
Participants in a conversation constantly shape the flow of the interaction through their choices. 
In psychological crisis counseling conversations, 
where counselors  support individuals in mental distress, 
these choices arise in uniquely complex and high-stakes circumstances,
and are reflected in rich conversational dynamics 
\cite{sacks_lectures_1992}.
As such, 
counseling is a valuable context 
for computationally modeling conversational behavior 
\cite{atkins_scaling_2014,althoff_large-scale_2016,perez-rosas_analyzing_2018,zhang_finding_2019}.
Modeling the conversational 
choices
 of counselors in
  this endeavor is an important step towards better supporting them.

\begin{figure}
\includegraphics[width=.47\textwidth]{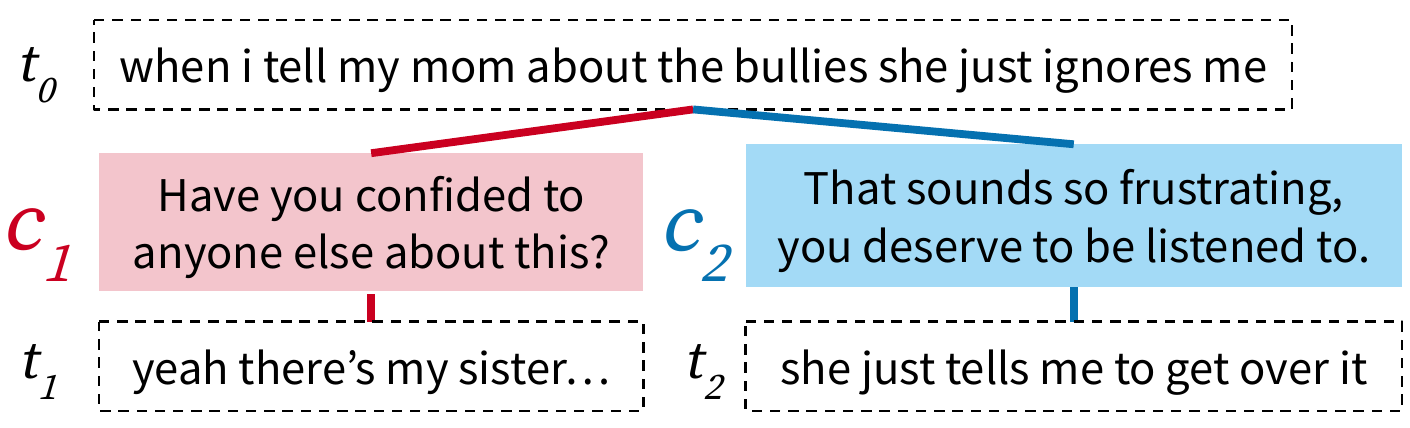}
\caption{Two possible exchanges in a counseling conversation, 
illustrating key objectives that a counselor must balance:
\fwdmessage aims to \textit{advance} the conversation towards a discussion of possible confidants;
\bkmessage aims to \textit{address} the emotion underlying the preceding utterance.
}
\label{fig:fig1}
\end{figure}

Counselors are driven by several objectives that serve the broader goal of helping the individual in distress;
two key objectives are 
exemplified
 in Figure \ref{fig:fig1}.\footnote{These examples are derived from material used to train counselors in our particular setting, detailed in Section \ref{sec:data}.} 
The counselor must \textit{advance} a conversation towards a calmer state where the individual is better equipped to cope with their situation \cite{mishara_which_2007,sandoval_crisis_2009}:
in \fwdmessage, the counselor prompts the individual to brainstorm options for social support.
The counselor must also 
empathetically
 \textit{address} what was already said, ``coming to an empathic understanding'' of the individual \cite{rogers_necessary_1957,hill_clientcentered_2000}:
in \bkmessage, the counselor 
validates feelings that the individual has just shared.
Balancing both objectives is often challenging, 
and 
overshooting in one direction
 can be detrimental
to the conversation.
A counselor who leans too much on advancing \emph{forwards}
could rush the conversation at the expense of establishing an empathetic connection;
 a counselor who leans 
too much
\emph{backwards},
on addressing 
what was already said,
 may fail to make any progress.

In this work, we develop a method to examine counselor behaviors as they relate to this balancing challenge. 
We quantify the relative extent to which an
utterance is aimed at advancing the conversation, versus addressing existing content.
We thus map each utterance onto a continuous backwards-forwards axis which models the balance of these objectives, and refer to an utterance's position on this axis as its \textit{orientation}.

At an intuitive level, our approach considers the range of content that is  expected to follow or precede a particular utterance. 
For an utterance like \fwdmessage that aims to advance the conversation towards an intended target, we would expect a narrow range of appropriate replies, concentrated around that target (e.g., suggestions of possible confidants).
We would likewise expect an utterance like \bkmessage that aims to address a previously-discussed situation to only be an appropriate reply for a narrow range of possible utterances, concentrated around that specific type of situation (e.g., disclosures of negative feelings).
 Starting from this intuition, we
  develop an unsupervised method 
  to quantify and compare these expected forwards and backwards ranges for any utterance, yielding our orientation measure.

Using this measure, we characterize counselor behaviors in a large collection of text-message conversations from a crisis counseling service,
which we accessed in collaboration with the service and with
the participants' consent.
We show how orientation meaningfully distinguishes between key conversational strategies that counselors are taught during their training.
We also show that our measure tracks a conversation's progress and can signal its effectiveness, highlighting the importance of balancing the advancing and addressing objectives, and laying the basis for future inquiries in establishing potential causal effects.

In summary, we develop an unsupervised methodology that captures how counselors balance the conversational objectives of advancing and addressing (Section \ref{sec:method}), apply and validate it in a large dataset of counseling conversations (Section \ref{sec:validation}),
and use it to investigate the relation between a counselor's conversational behavior and their  effectiveness (Section \ref{sec:analysis}).
While our method is motivated by a salient challenge in counseling,
we expect similar balancing problems to recur in other conversational settings where participants must carefully direct the flow of the interaction, such as court trials
and debates (Section~\ref{sec:discussion}).

\section{Setting: Counseling Conversations}
\label{sec:data}
We develop our method
in the context of 
Crisis Text Line,
a 
crisis counseling platform
which provides a free 24/7 service for anyone in 
mental crisis---henceforth \textit{texters}---to have one-on-one 
conversations via text message with affiliated counselors. 
We accessed a version of this collection,
with over 1.5 million conversations, in collaboration with the platform and with the consent of the participants. 
The data was scrubbed of personally identifiable information by the platform.\footnote{The data can be accessed by applying at \footnotesize{\url{https://www.crisistextline.org/data-philosophy/data-fellows/}}. The extensive ethical and privacy considerations, and policies accordingly implemented by the platform, are detailed in \citet{pisani_protecting_2019}.}
These conversations are quite substantive, averaging 25 messages with 29 and 24 words per counselor and texter message, respectively.

In each conversation, a crisis counselor's high-level goal is to guide the texter towards a calmer mental state.
In service of this goal, all counselors first complete 30 hours of training provided by the platform,
which draws on past literature in counseling to recommend best practices
and conversational strategies.
The first author also completed the training to 
gain familiarity with the domain.

While the platform offers guidance to counselors, their task is inevitably open-ended, given the emotional complexity of
crisis situations.
As such, the counselors are motivated by an explicit goal that structures the interaction,
but they face a challenging flexibility in choosing how to act.
\section{Background and Related Work}
\label{sec:idea}

We now describe the conversational challenge 
of balancing between advancing the conversation forwards or addressing what was previously said.
Our description of the challenge and our computational approach to studying it
are informed by literature in counseling, 
on the platform's training material and on informal interviews with its staff.

\xhdr{A conversational balance}
A crisis counselor must fulfill multiple objectives in their broader goal of helping a texter.
One objective is guiding the texter through their initial distress to a calmer mental state \cite{mishara_which_2007,sandoval_crisis_2009}, as in Figure \ref{fig:fig1}, \fwdmessage.
Various strategies that  aim to facilitate this \textit{advancing} process  are taught to counselors during training:
for instance, a counselor may prompt a texter to identify a goal or coping mechanism \cite{rollnick_what_1995}.
As such, 
they attempt to move
the conversation \textit{forwards}, towards its eventual resolution.

The counselor must also engage with the texter's concerns \cite{rogers_necessary_1957,hill_clientcentered_2000}, as in \bkmessage,
via strategies that empathetically \textit{address} what the texter has 
already shared \cite{rollnick_what_1995,weger_active_2010,bodie_role_2015}.
For instance, counselors are taught to \textit{reflect}, i.e., reframe a texter's previous message to 
convey understanding,
or draw on what was said to affirm the texter's positive qualities. 
In doing so, the counselor looks \textit{backwards} in the conversation.

Past work has posited the benefits of mixing between strategies that aim at either objective \cite{mishara_which_2007}.
However, as the training acknowledges, striking this balance is challenging.
Overzealously seeking to advance could cut short 
the process of establishing an empathetic connection.
Conversely, focusing on the conversation's past may 
not help with
eventual problem solving \cite{bodie_role_2015}, and risks stalling it.
A texter may start to counterproductively rehash 
or \textit{ruminate} on their concerns \cite{nolen-hoeksema_rethinking_2008,jones_over_2009};
indeed, prior psychological work has highlighted the thin line between productive reflection and rumination 
\cite{rose_prospective_2007,landphair_more_2012}.

\xhdr{Orientation}
To examine this balancing dynamic, we model the choices that counselors make at each turn in a conversation.
Our approach is to derive a continuous axis spanned by advancing and addressing.
We refer to an utterance's position on this axis,
representing the relative extent to which it aims at either objective, as its \textit{orientation} \orientation{}.
We interpret a \textit{forwards-oriented} utterance with positive \orientation{} as aiming to advance the conversation,
and a \textit{backwards-oriented} utterance with negative \orientation{} as aiming to address what was previously brought up. 
In the middle, the axis reflects the graded way in which a counselor can balance between aims---for instance,
using something the texter has previously said to help motivate a problem-solving strategy.

\xhdr{Related characterizations}
While we develop orientation 
to model 
a dynamic in counseling,
we view it as a complement to other characterizations of conversational behaviors
in varied settings.

Prior work has similarly 
considered how utterances relate to the preceding and subsequent discourse \cite{webber_computational_2001}.
Frameworks like centering theory \cite{grosz_centering_1995} aim at identifying referenced entities, 
while we aim to more abstractly model interlocutor choices.
Past work has also examined how interlocutors mediate 
a conversation's trajectory through
taking or ceding control 
\cite{walker_mixed_1990}
or shifting 
topic 
\cite{nguyen_modeling_2014};
\citet{althoff_large-scale_2016} considers the rate at which counselors in 
our setting
advance across stages of a conversation. 
While these actions can be construed as forwards-oriented, we focus more on the interplay between forwards- and backwards-oriented actions.
A counselor's objectives may also cut across these concepts:
for instance, the training stresses the need for empathetic reflecting across all stages
and topics.
Orientation also complements prior work on dialogue acts, which 
consider various roles that utterances play in discourse \cite{mann_rhetorical_1988,core_coding_1997,ritter_unsupervised_2010,bracewell_identification_2012,rosenthal_i_2015,prabhakaran_detecting_2018,wang_persuasion_2019}.
In counseling settings, 
such approaches have highlighted strategies like reflection and question-asking \cite{houck_motivational_2008,gaume_counselor_2010,atkins_scaling_2014,can_dialog_2015,tanana_comparison_2016,perez-rosas_understanding_2017,perez-rosas_analyzing_2018,park_conversation_2019,lee_identifying_2019,cao_observing_2019}.
Instead of modeling a particular taxonomy of actions, we model how counselors balance among the 
underlying objectives;
we later relate orientation to these strategies (Section~\ref{sec:validation}).
Most of these approaches 
use
annotations or predefined labeling schemes, while our method is unsupervised.

\section{Measuring Orientation}
\label{sec:method}
We now describe our method to measure orientation,
discussing our approach at a high level before elaborating on the particular operationalization.
The code implementing our approach is distributed as part of the ConvoKit library \cite{chang_convokit_2020}, at \url{http://convokit.cornell.edu}.

\subsection{High-Level Sketch}
Orientation compares the extent to which an utterance aims to advance the conversation forwards with the extent to which it looks backwards. 
Thus, we must somehow quantify how the utterance relates to the subsequent and 
preceding interaction.

\xhdr{Naive attempt: direct comparison}
As a natural starting point, we may opt for a similarity-based approach:
an utterance that aims to address its preceding utterance,
or \textit{predecessor},
should be similar to it;
an utterance that aims to advance the conversation should be similar to the reply that it prompts.
In practice, having to make these direct comparisons is limiting:
an automated system 
 could not
  characterize an utterance in an ongoing conversation by comparing it to
  a reply it has yet to receive.

This approach also has important conceptual faults.
First, addressing preceding content in a conversation is different from recapitulating it.
For instance, counselors are instructed to \textit{reframe} rather than outright restate a texter's message, 
as in Figure~\ref{fig:fig1}, \bkmessage.
Likewise, counselors need not advance the conversation by declaring something for the texter to simply repeat;
rather than giving specific recommendations, counselors are instructed to prompt the texters to come up with coping strategies on their own, 
as in \fwdmessage.
Further, 
texters are not bound to the relatively formal linguistic style counselors must maintain, 
resulting in 
clear
 lexical differences.
Measuring orientation is hence a distinct task from measuring similarity.

Second, an utterance's \textit{intent} to advance need not actually be realized.
A counselor's cues may be rebuffed or misunderstood \cite{schegloff_sources_1987,thomas_cross-cultural_1983}: 
a texter could respond to \fwdmessage by continuing to articulate their problem with \bktxtmessage.
Likewise, a counselor may intend to address a texter's concerns but misinterpret them.
To model the balance in objectives that a counselor is aiming for,
our characterization of an utterance cannot be contingent on its actual reply and predecessor.

\begin{figure}
\includegraphics[width=.47\textwidth]{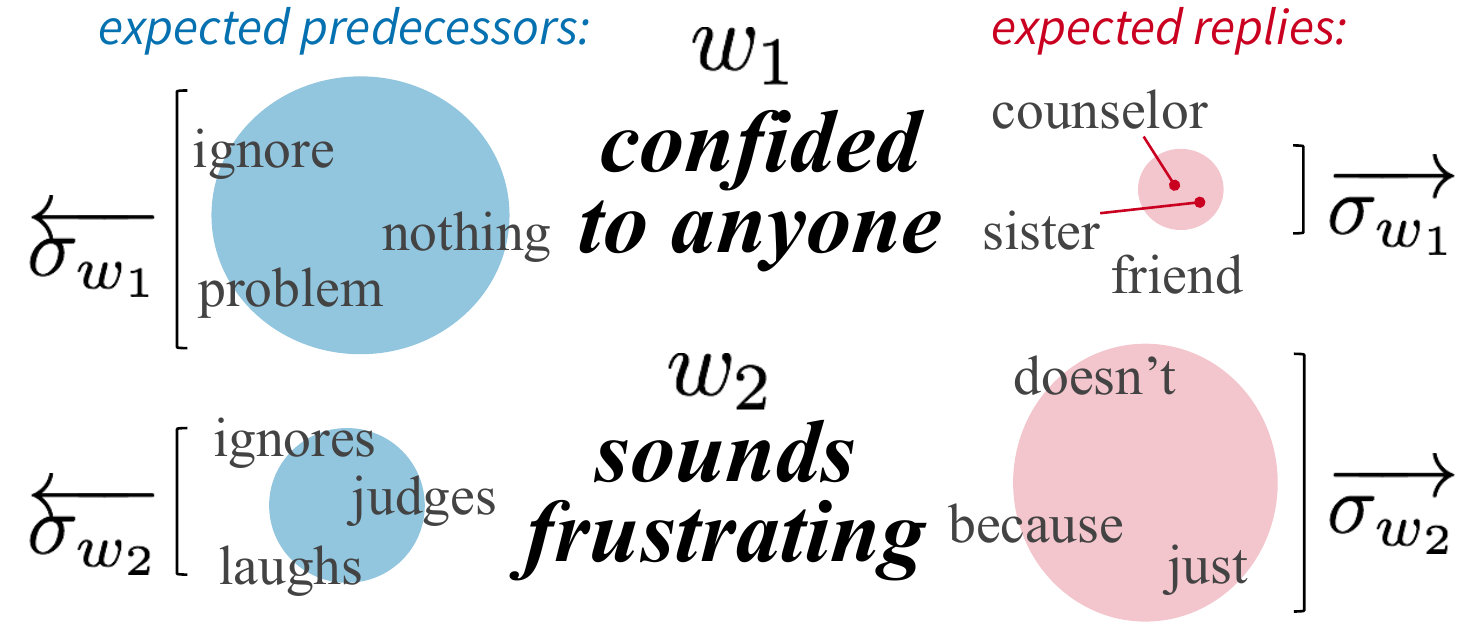}
\caption{
Words representative of replies and predecessors for utterances with two example phrasings, as observed in training data.
Top row: observed replies to utterances with \fwdword span a narrower range than observed predecessors (relative sizes of red and blue circles);  \fwdword thus has smaller \textit{forwards-range} \rightattn{\fwdword} than \textit{backwards-range} \leftattn{\fwdword} (i.e., it is forwards-oriented, $\orientation{\fwdword}>0$).
Bottom row: observed predecessors to utterances with \bkword span a narrower range than replies; \bkword thus has smaller \leftattn{\bkword} than \rightattn{\bkword} (i.e., it is backwards-oriented $\orientation{\bkword}<0$). 
}
\label{fig:fig2}
\end{figure}

\xhdr{Our approach: characterizing expectations}
We instead consider the range of replies we might \textit{expect} an utterance to receive,
or the range of predecessors that it might follow.
Intuitively, an utterance with a narrow range of appropriate replies aims to direct the conversation towards a particular target,
moreso than an utterance whose appropriate replies span a broader range.\footnote{Consider leading versus open-ended questions. When people ask leading questions, they intend to direct the interaction towards specific answers they have in mind; when people ask open-ended questions, they are more open to what answers they receive and
where the interaction is headed.}
Likewise, an utterance that is an appropriate reply to only 
a narrow range of possible predecessors aims to address a particular situation.
We draw on unlabeled data of past conversations to form our expectations of these ranges,
and build up our characterizations of utterances from their constituent \textit{phrasings}, e.g., words or dependency-parse arcs.

The intuition for our approach is sketched in Figure \ref{fig:fig2}.
From our data, we observe that utterances containing
\egword{confided to anyone} generally elicited replies about potential confidants (e.g., \egword{sister}, \egword{friend}),
while the replies that followed utterances with \egword{sounds frustrating} span a broader, less well-defined range.
As such, we have a stronger expectation of what a reply prompted by a \textit{new} utterance with \egword{confided to anyone} might contain
than
a reply to a new utterance with \egword{sounds frustrating}.
More generally, for each phrasing \word, 
we quantify the strength of our expectations of its potential replies 
by measuring the range spanned by the replies it has already received in the data,
which we refer to as its \textit{forwards-range} \rightattn{\word}. 
We would say that \egword{confided to anyone} has a smaller \rightattn{\word} than \egword{sounds frustrating}, meaning that its observed replies were more narrowly concentrated;
this is represented 
as the relative size of the red regions on the right side of Figure \ref{fig:fig2}.

In the other direction, we observe in our data that \egword{sounds frustrating} generally followed descriptions of frustrating situations (e.g., \egword{ignores}, \egword{judges}),
while the range of predecessors to \egword{confided to anyone} is broader.
We thus have a stronger expectation of the types of situations that new utterances with \egword{sounds frustrating} would respond to,
compared to new utterances with \egword{confided to anyone}.
For a phrasing \word, we quantify the strength of our expectations of its potential predecessors by measuring its \textit{backwards-range} \leftattn{\word}, 
spanned by the predecessors we've observed.
As such, \egword{sounds~frustrating} has a smaller \leftattn{\word} than \egword{confided to anyone},
corresponding to the relative size of the blue regions on the left side of Figure \ref{fig:fig2}.

The relative strengths of our expectations in either direction 
then indicate the balance of objectives. 
If we have a stronger expectation of \word's replies than of its predecessors---i.e., smaller \rightattn{\word} than \leftattn{\word}---we 
would infer that utterances with \word aim to advance the conversation towards a targeted reply more than they aim to address a particular situation.
Conversely, if we have stronger expectations of \word's predecessors---i.e., smaller \leftattn{\word}---we
would infer that utterances with \word aim to address the preceding interaction, rather than trying to drive the conversation towards some target.

We thus measure orientation by comparing a phrasing's forwards- and backwards-range. 
The expectation-based approach allows us to circumvent the shortcomings of a direct comparison;
we may interpret it as modeling a counselor's \textit{intent} in advancing and addressing at each utterance \cite{moore_planning_1993,zhang_asking_2017}.

\subsection{Operationalization}
\label{sec:method_op}

\begin{figure}
\includegraphics[width=0.48\textwidth]{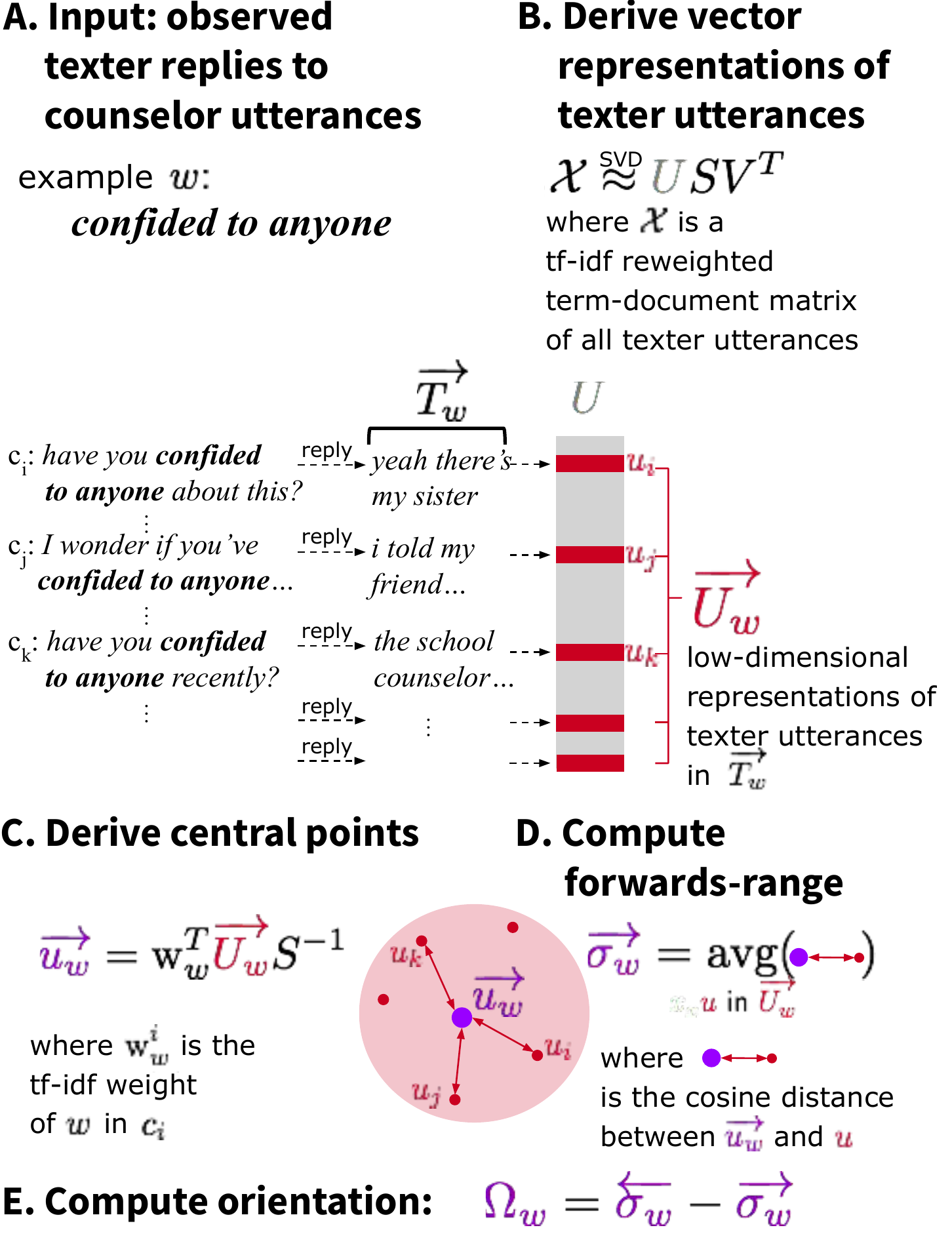}
\caption{Outline of steps to compute orientation \orientation{\word} of phrasing \word, as described in Section \ref{sec:method_op}.
Panels A-D show the procedure for computing forwards-range \rightattn{\word}; 
the procedure for backwards-range \leftattn{\word} is similar.
}
\label{fig:method}
\end{figure}

We now detail the steps of our method, which are outlined in Figure \ref{fig:method}.
Formally, our input consists of a set of utterances from counselors $\{\ccmessage\}$, and a set of utterances from texters $\{\txmessage\}$, which we've observed in a dataset of conversations (Figure \ref{fig:method}A).
We note that each texter utterance can be a reply to, or a predecessor of, a counselor utterance (or both). 
We use this unlabeled ``training data'' to measure the forwards-range \rightattn{\word}, 
the backwards-range \leftattn{\word} (Figures \ref{fig:method}B-D),
and hence the orientation \wordor of each phrasing \word used by counselors (Figure \ref{fig:method}E).
We then aggregate to an utterance-level measure.

For each counselor phrasing \word, 
let \rightmessages{\word} denote the subset of texter utterances which are replies to counselor utterances containing \word (Figure \ref{fig:method}A).
As described above, the forwards-range \rightattn{\word} quantifies the 
spread among elements of \rightmessages{\word};
we measure this by deriving vector representations of these utterances \righttxsub{\word} (Figure \ref{fig:method}B, detailed below), 
and then comparing each vector in  \righttxsub{\word}  to a central reference point \fwdcentre{\word} (Figures \ref{fig:method}C and \ref{fig:method}D).\footnote{Using a central reference point to calculate the forwards-range, as opposed to directly computing pairwise similarities among replies in \righttxsub{\word}, 
allows us to account for the context of \word in the utterances that prompted these replies (via tf-idf weighting, as subsequently discussed). 
}
Likewise, \leftattn{\word} quantifies the similarity among elements of \leftmessages{\word}, the set of predecessors to counselor utterances with \word;
we compute \leftattn{\word} by comparing each corresponding vector in \lefttxsub{\word}  to a central  point \bkcentre{\word}.

\xhdr{Deriving vector representations (Figure \ref{fig:method}B)}
To obtain vectors for each texter utterance, we construct \txmtx, a tf-idf reweighted term-document matrix 
where rows represent texter utterances and columns represent phrasings used by texters.
To ensure that we go beyond lexical matches 
and
 capture conceptual 
classes (e.g., possible confidants, frustrating situations),
we use singular value decomposition to get 
$\txmtx \approx \txrepr SV^T$.
Each row of \txrepr is a vector representation \txvect of utterance \txmessage in the induced low-dimensional space \latentspace.
\righttxsub{\word} then consists of the corresponding subset of rows of \txrepr (highlighted in Figure \ref{fig:method}B).

\xhdr{Deriving central points (Figure \ref{fig:method}C)}
For each \word, we take 
the
central point
\fwdcentre{\word} to be a weighted average of vectors in \righttxsub{\word}. 
Intuitively, a texter utterance \txmessage with vector \txvect should have a larger contribution to \fwdcentre{\word} 
if \word is more prominent in the counselor utterance \ccmessage that preceded it.
We let \wordrepr{i}{\word} denote the normalized tf-idf weight of \word in \ccmessage, 
and use \wordrepr{i}{\word} as the weight of the corresponding vector \txvect.
To properly map the resultant weighted sum $\sum \wordrepr{i}{\word} \txvect$ into \latentspace, we 
divide each dimension by the corresponding singular value in $S$. As such, if \wordreprs{\word} is a vector of weights \wordrepr{i}{\word}, we can calculate the central point \fwdcentre{\word} of \righttxsub{\word} as $\fwdcentre{\word} = \wordreprs{\word}^T \righttxsub{\word} S^{-1}$.
In the other direction, we likewise compute $\bkcentre{\word} = \wordreprs{\word}^T \lefttxsub{\word} S^{-1}$.

\xhdr{Forwards- and backwards-ranges (Figure \ref{fig:method}D)}
We take the forwards-range \rightattn{\word} of \word to be the average cosine distance from each vector in \righttxsub{\word} to the center point \fwdcentre{\word}. 
Likewise, we take 
\leftattn{\word} as the average distance from each vector in \lefttxsub{\word} to \bkcentre{\word}.

\xhdr{Phrasing-level orientation (Figure \ref{fig:method}E)}
Importantly, since we've computed
the forwards- and backwards-ranges
\rightattn{\word} and \leftattn{\word}
using distances in the same space \latentspace,
their values are comparable. 
We 
then 
 compute the orientation of \word as their difference: 
$\orientation{\word} = \leftattn{\word} - \rightattn{\word}$.

\xhdr{Utterance-level orientation}
To compute the orientation of an utterance \ccmessage, 
we first compute the orientation of each sentence in \ccmessage as the tf-idf weighted average \orientation{\word} of its constitutent phrasings.
Note that a multi-sentence utterance can orient in \textit{both} directions---e.g., 
a counselor could concatenate \bkmessage and \fwdmessage from Figure \ref{fig:fig1} in a single utterance, addressing the texter's previous utterance before moving ahead. 
To model this heterogeneity, we consider both the minimum and maximum sentence-orientations in an utterance:
\minorientation{} captures the extent to which the utterance looks backwards,
while \maxorientation{} captures the extent to which it aims to advance forwards.

\section{Application to Counseling Data}
\label{sec:validation}
We apply our method to characterize messages 
from 
crisis
counselors on the platform. 
We compute the orientations of the phrasings they use,
represented as dependency-parse arcs.
We use a training set of 351,935 texter and counselor messages each,
from a random sample of conversations omitted in subsequent analyses.\footnote{Further implementation details are listed in the appendix.}
Table~\ref{tab:examples} shows representative phrasings and sentences of different orientations.\footnote{Example sentences are derived from real sentences in the data, and modified for readability. 
The examples were chosen to reflect common situations in the data,
and were vetted by the platform to ensure the privacy of counselors and texters.}
Around two-thirds of phrasings and sentences have $\orientation{}\!<\!\!0$, 
echoing the importance of addressing the texter's previous remarks.

In what follows, we analyze counselor behaviors in terms of orientation, and illustrate how the measure can be useful for examining conversations.
We start by validating our method via two complementary approaches.
In a subset of sentences manually annotated with the counseling strategies they exhibit, we show that orientation meaningfully reflects these strategies (Section \ref{sec:validation_check}).
At a larger scale, we show that the orientation of utterances over the course of a conversation aligns with domain knowledge about counseling conversation structure (Section \ref{sec:validation_struct}).
We also find that other measures for characterizing utterances are not as rich as orientation in capturing counseling strategies and conversation structure (Section \ref{sec:validation_alt}).
Finally, we show that a counselor's orientation in a conversation is tied to indicators of their effectiveness in helping the texter (Section \ref{sec:analysis}).

\begin{table*}[]
\small
\begin{tabular}{lll}
\hline
\textbf{Orientation}                                                   & \textbf{Example phrasings}                                                                                                                                  & \textbf{Example sentences}                                                                                                                                                                                                                                                                                            \\ \hline
\begin{tabular}[c]{@{}l@{}}Backwards-\\ oriented \\ (bottom 25\%)\end{tabular} & \begin{tabular}[c]{@{}l@{}@{}@{}}sounds frustrating, totally normal, \\ great ways, on {[}your{]} plate, \\ be overwhelming, sometimes feel \\ frightening, on top [of] \\ been struggling, feeling alone\end{tabular} & \begin{tabular}[c]{@{}l@{}@{}@{}@{}}You have a lot of things on your plate, between family\\ \qquad and financial problems. \textbf{[reflection]}\\ It's totally normal to feel lonely when you have \\ \qquad no one to talk to. \textbf{[reflection]}\\ Those are  great ways to  improve the relationship. \textbf{[affirmation]} \end{tabular} \\ \hline 
(middle 25\%)
              & \begin{tabular}[c]{@{}l@{}@{}}happened {[}to{]} make, \\ mean {[}when you{]} say,\\ is that, you recognized, source of \\ the moment, are brave\end{tabular}                    & \begin{tabular}[c]{@{}l@{}}Has anything happened  to make you anxious? \textbf{[exploration]}\\ It's good you recognized the need to reach out.  \textbf{[affirmation]}\\ 
              Can you tell me what you mean when you say \\ \qquad you're giving up? \textbf{[risk assessment]} \end{tabular}                            \\ \hline
\begin{tabular}[c]{@{}l@{}@{}}Forwards-\\ oriented \\ (top 25\%) \end{tabular}  & \begin{tabular}[c]{@{}l@{}}plan for, confided {[}to{]} anyone,\\ usually do, has helped, \\ been talking, best support \\ have considered, any activities \end{tabular}                & \begin{tabular}[c]{@{}l@{}@{}}Can you think of anything that has helped when \\ \qquad you've been stressed before? \textbf{[problem solving]}\\
 I want to be the best support for you today. \textbf{[problem solving]}  \\ We've been talking for a while now, how  do you feel? \textbf{[closing]}\end{tabular}                                                                           \\ \hline
\end{tabular}
\caption{Example phrasings and sentences with labeled strategies 
from crisis counselors' messages,
at varying orientations: backwards-oriented (from the bottom 25\% of \orientation{}), 
middle, 
and forwards-oriented (from top 25\%). 
}
\label{tab:examples}
\end{table*}

\subsection{Validation: Counseling Strategies}
\label{sec:validation_check}

Even though it is computed without the guidance of any annotations,
we expect orientation to meaningfully reflect 
strategies for advancing or addressing that 
crisis
counselors 
are taught.
The first author hand-labeled 400 randomly-selected sentences with a set of pre-defined strategies derived from techniques highlighted in the training material.
We note example sentences in Table~\ref{tab:examples} which exemplify each strategy, and provide more extensive descriptions in the appendix.

Figure~\ref{fig:val_fig}A depicts the distributions of orientations across each label,
sorted from most backwards- to most forwards-oriented.
We find that the relative orientation of different strategies corroborates their intent as described in the literature.
Statements \textbf{reflecting} or \textbf{affirming} what the texter has said to check understanding or convey empathy
(characterized by phrasings like \egword{totally normal}) tend to be backwards-oriented; 
statements prompting the texter to advance towards \textbf{problem-solving}
(e.g., \egword{[what] has helped}) 
are more forwards-oriented.
\textbf{Exploratory} queries for more information on what the texter has already said 
(e.g., \egword{happened to make}) tend to have middling orientation (around 0). 
The standard deviation of orientations over messages within most of the labels is significantly lower than across labels
(bootstrapped $p\!<\!.05$, solid circles),
showing that orientation yields interpretable groupings of messages in terms of important counseling strategies.

The measure also offers complementary information.
For instance, we find sentences that aren't accounted for by pre-defined  labels, but still map to interpretable orientations,
such as backwards-oriented examples assuaging texter concerns about the platform being a safe space to self-disclose.

\subsection{Validation: Conversation Structure}
\label{sec:validation_struct}
We also show that orientation tracks with the structure of 
crisis
counseling conversations as described in the training material.
Following \citet{althoff_large-scale_2016}, we divide each conversation with at least ten counselor messages into five equal-sized segments and average \maxorientation{} and \minorientation{} over messages in each segment. 

\begin{figure}[]
\includegraphics[width=.48\textwidth]{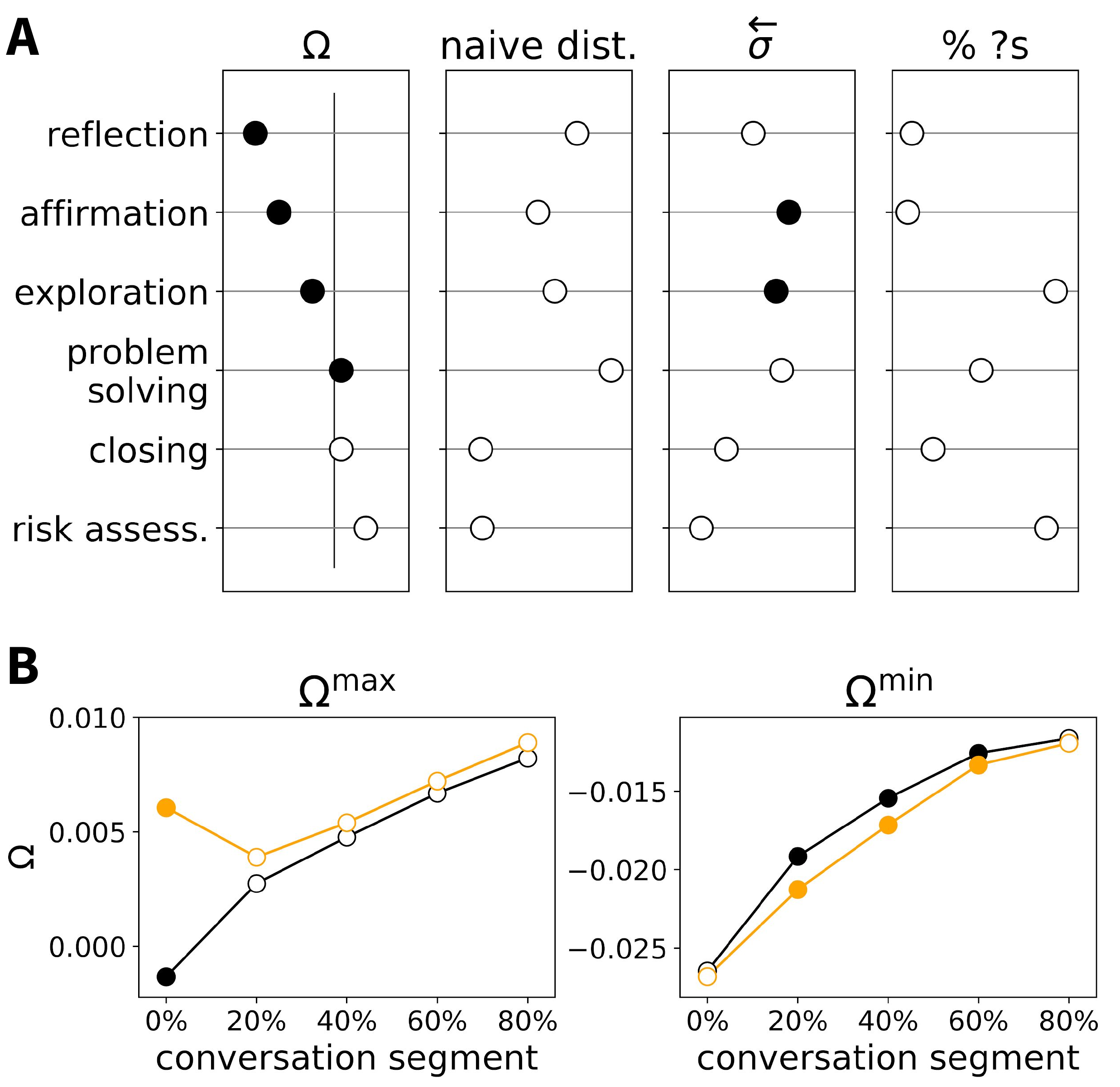}
\caption{
Validating the orientation measure and comparing to alternatives.
\textbf{A} Leftmost: Mean \orientation{} per counseling strategy label (vertical line denotes $\orientation{}\!=\!0$). Next three: same for other measures. \textbf{B}: Mean \maxorientation{} and \minorientation{} per segment for risk-assessed (orange) and non-risk-assessed (black) conversations. 
Both: Solid circles indicate statistically significant differences 
(Wilcoxon $p\!<\!0.01$, comparing within-counselor). 
}
\label{fig:val_fig}
\end{figure}

Figure~\ref{fig:val_fig}B (black lines) shows that 
over the course of a conversation, messages tend to get more forwards-oriented (higher \maxorientation{} and \minorientation{}).
This matches a standard conversation structure taught in the training: 
addressing the texter's existing problems before advancing towards problem-solving.
While this correspondence holds in aggregate, 
orientation also captures complementary information to advancement through stages---e.g.,
while problem-solving, counselors may still address and affirm a texter's ideas (Table \ref{tab:examples}, row 3).

We also consider a subset of conversations
where we expect a different trajectory:
for potentially suicidal texters,
the training directs counselors to immediately start a process of \textbf{risk assessment} in which 
actively prompting 
the texter to disclose their level of suicidal ideation takes precedence over other objectives.
As such, we expect more forwards-oriented messages at the starts of conversations involving such texters.
Indeed, in the 30\% of conversations which are risk-assessed, 
we find significantly larger \maxorientation{} in the first segment
(Figure \ref{fig:val_fig}B, orange line; Wilcoxon $p\!<\!0.01$ in the first stage, comparing within-counselor). 
\minorientation{} is \textit{smaller} at each stage,
suggesting that 
counselors balance actively prompting these critical disclosures
with addressing them. 

\subsection{Alternative Operationalizations}
\label{sec:validation_alt}

We compare orientation to other methods for capturing a counselor's balancing decisions:

\xhdr{Naive distance} 
We conside the naive approach in Section \ref{sec:method}, 
taking a difference in cosine distances between tf-idf representations of a message and its reply, 
and a message and its predecessor.

\xhdr{Backwards-range}
We consider just the message's backwards-range.
For each sentence we take tf-idf weighted averages of component \leftattn{w} 
and take minimum \leftattn{} for each message.\footnote{We get qualitatively similar results with maximum \rightattn{}.}

\xhdr{Question-asking}
We consider whether the message has a question.
This was used in \citet{walker_mixed_1990} as a signal of taking control, 
which could be construed as forwards-oriented.

Within-label standard deviations of each alternative measure are generally not significantly smaller than across-label (Figure \ref{fig:val_fig}A), 
indicating that these measures are poorer reflections of the counseling strategies.
Label rankings under the measures are also arguably less intuitive.
For instance, reflection statements have relatively large (naive) cosine distance from their predecessors.
Indeed, the training encourages counselors to \textit{process} rather than simply restate the texter's words.

These measures also track with the conversation's progress differently---notably, 
none of them distinguish the initial dynamics of risk-assessed conversations as reflected in \maxorientation{} (see appendix).
\subsection{Relation to Conversational Effectiveness}
\label{sec:analysis}
Past work on counseling has extensively discussed the virtues of addressing a client's situation 
\cite{rogers_necessary_1957,hill_clientcentered_2000}.
Some studies also suggest that 
accounting for \textit{both} addressing and advancing is important,
such that effective counselors manage to mix backwards- and forwards-oriented actions \cite{mishara_which_2007}.

We use orientation to examine how these strategies are tied to conversational effectiveness 
in crisis counseling
at a larger scale,
using our measures to provide
a unified view of advancing and addressing.
To derive simple conversation-level measures,
we average 
\maxorientation{} and \minorientation{} over each counselor message in a conversation.

Adjudicating counseling conversation quality is known to be difficult \cite{tracey_expertise_2014}. 
As a starting point, we relate our conversation-level measures to two complementary indicators of a conversation's effectiveness:\footnote{We perform all subsequent analyses on a subset of 234,433 conversations, detailed in the appendix.}

\begin{figure}
\includegraphics[width=.48\textwidth]{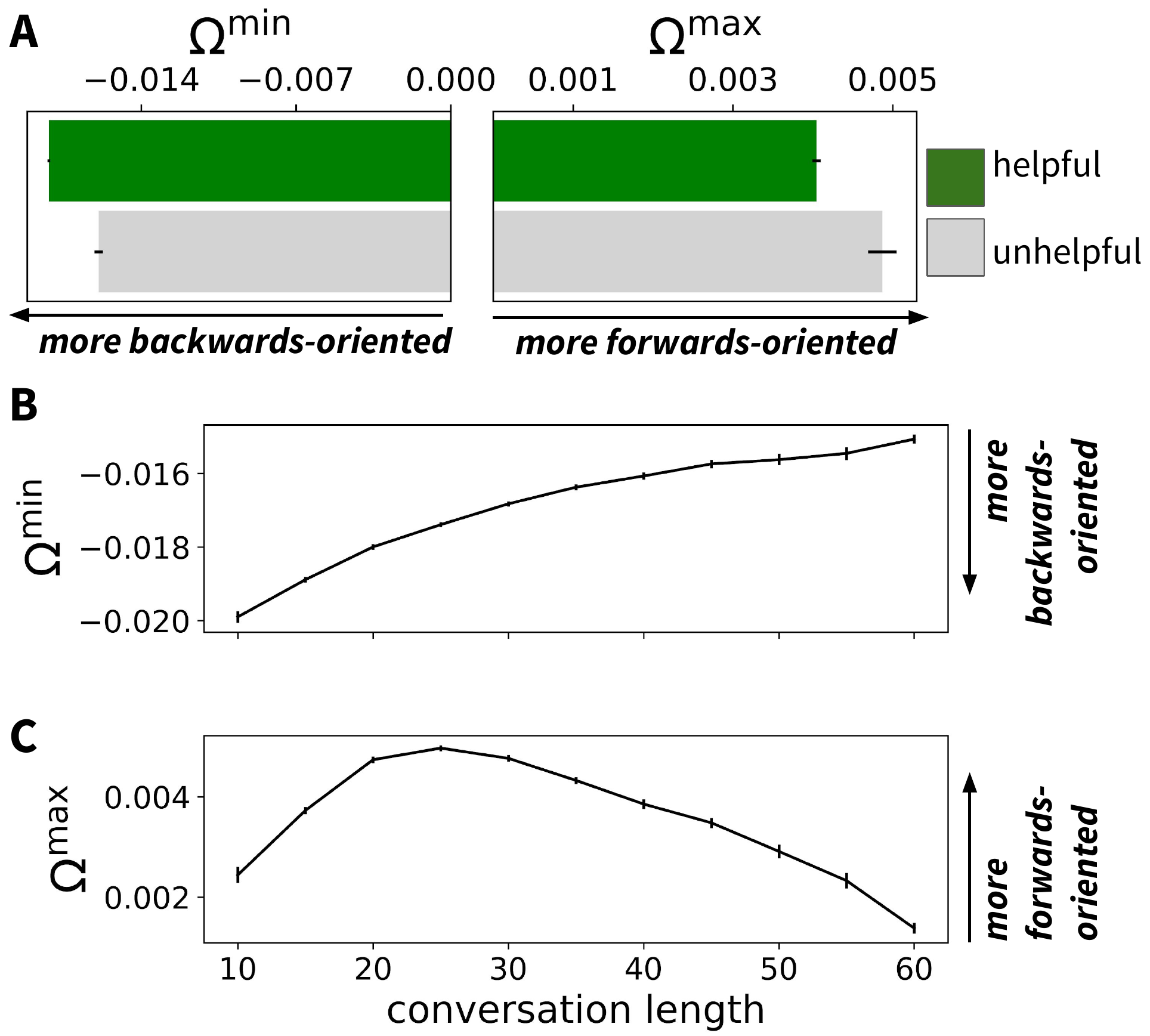}
\caption{
Relation between orientation and conversational effectiveness.
\textbf{A}: Mean \minorientation{} and \maxorientation{} in conversations  rated as helpful (green) or unhelpful (grey)
(macroaveraged per conversation). 
Differences in both measures are significant (Mann Whitney U test $p < 0.001$).
\textbf{B}, \textbf{C}: Mean \minorientation{} and \maxorientation{} of conversations with varying lengths
 (in \# of messages).
Both plots: Error bars show 95\% bootstrapped confidence intervals.
}
\label{fig:analysis}
\end{figure}

\xhdrit{Perceived helpfulness}
We consider responses from a post-conversation survey asking the texter whether
the conversation was helpful, following \citet{althoff_large-scale_2016}.
Out of the 26\% of conversations with a response, 89\% were rated as helpful.\footnote{We note that this indicator is limited by important factors such as the selection bias in respondents.}

\xhdrit{Conversation length}
We consider a conversation's length as a simple indicator of the pace of its progress:
short conversations may rush the texter,
while prolonged conversations could suggest stalling and could even 
demoralize the counselor \cite{landphair_more_2012}.\footnote{As the training material notes, conversation length and texter perception may signal complementary or even conflicting information about a texter's experience of a conversation and its effectiveness: ``Some texters resist the end of the conversation. They ruminate [...] causing the conversation to drag on without any progress.''}

Figure \ref{fig:analysis}A compares 
\minorientation{} and \maxorientation{} in conversations rated as helpful and unhelpful by texters. 
Both measures are significantly \textit{smaller} in conversations perceived as helpful,
suggesting that texters have a better impression of relatively \textit{backwards-oriented} interactions where the counselor is inclined towards addressing their situation.
As such, this result echoes past findings relating addressing to effectiveness.

Figure \ref{fig:analysis}B compares \minorientation{} in conversations of varying lengths, showing that \minorientation{} \textit{increases} with length, such that counselors 
exhibit less propensity for addressing in longer conversations.
Anecdotal observations cited in interviews with the platform's staff suggest one interpretation: 
conversations in which a
texter feels their concerns were not satisfactorily addressed may be prolonged when they circle back to revisit these concerns.

Figure \ref{fig:analysis}C relates \maxorientation{} to conversation length.
We find that \maxorientation{} is smaller in the lengthiest conversations, suggesting that such prolonged interactions may be stalled by a weaker impulse to advance forwards.
Extremely short conversations have smaller \maxorientation{} as well, such that premature endings may also reflect issues in advancing.
As such, we add credence to the previously-posited benefits of mixing addressing and advancing: forwards-oriented actions may be tied to making progress, while a weaker propensity to advance may signal a suboptimal pace.

\xhdr{Counselor-level analysis}
These findings could reflect various confounds---for instance, a counselor's choice of orientation may have no bearing on the rating they receive from a particularly difficult texter.
To address this, we compute similar correspondences between orientation and our effectiveness indicators at the level of counselors rather than conversations; 
this analysis is detailed in the appendix.
Our conversation-level results are replicated under these controls.

\section{Discussion and Future Work}
\label{sec:discussion}
In this work, we sought to examine a key balance in crisis counseling conversations between advancing forwards and addressing what has already been said.
Realizing this balance is 
one of the many challenges that 
crisis
counselors must manage,
and modeling the actions they take in light of such challenges
could point to 
policies to better support them.
For instance, 
our method could assist human supervisors in monitoring the progress of ongoing conversations to detect instances of rushing or stalling,
or enable larger-scale analyses of conversational behaviors to inform how counselors are trained.
The unsupervised approach we propose 
could circumvent difficulties in getting large-scale annotations of such sensitive content.

Future work could bolster the measure's usefulness in several ways. 
Technical improvements like richer utterance representations could improve the measure's fidelity;
more sophisticated analyses could better capture the dynamic ways in which the balance of objectives is negotiated across many turns.
The preliminary explorations in Section \ref{sec:analysis} could also be extended to gauge the causal effects of counselors' behaviors \cite{kazdin_mediators_2007}.

We expect balancing problems to recur in conversational settings beyond crisis counseling, such as court proceedings, interviews, debates
and other mental health contexts like long-term therapy.
In these settings, individuals also make potentially consequential choices that span the backwards-forwards orientation axis, such as addressing previous arguments \cite{tan_winning_2016,zhang_conversational_2016} or asking leading questions \cite{leech_asking_2002}.
Our measure is designed to be 
broadly applicable,
requiring no domain-specific annotations; 
we provide exploratory output on justice utterances from the Supreme Court's oral arguments 
in the appendix
and release code implementing our approach 
 at \url{http://convokit.cornell.edu}
 to encourage experiments in other domains.
However, the method's efficacy in the present setting is likely boosted by the
relative uniformity of crisis counseling conversations;
and future work could aim to better accomodate settings with less structure and more linguistic variability.
With such improvements, it would be interesting to study other domains where interlocutors are 
faced with conversational challenges.
{
\xhdr{Acknowledgements}
We thank Jonathan P. Chang, Caleb Chiam, Liye Fu, Dan Jurafsky, Jack Hessel, and Lillian Lee for helpful conversations, and the anonymous reviewers for their thoughtful comments.
We also thank Ana Smith for collecting and processing the Supreme Court oral argument transcripts we used in the supplementary material.
This research, and the counseling service examined herein, would not have been possible without Crisis Text Line. We are particularly grateful to Robert Filbin, Christine Morrison, and Jaclyn Weiser for their valuable insights into the experiences of counselors and for their help with using the data.   The research has been supported in part by  NSF CAREER Award IIS1750615 and a Microsoft Research PhD Fellowship. The collaboration with Crisis Text Line was supported by the Robert Wood Johnson Foundation; the views expressed here do not necessarily reflect the views of the foundation.
}

\bibliography{orientation-acl-jz}
\bibliographystyle{acl_natbib}

\appendix

\label{sec:appendix}

\section{Appendices}

\subsection{Further Details About Methodology}
Here, we provide further details on our methodology for measuring orientation, to supplement the description in Section \ref{sec:method_op} and aid reproducibility.

Our aim in the first part of our methodology is to measure the orientation of phrasings \orientation{w}.
We would like to ensure that our measure is not skewed by the relative frequencies of phrasings, and take two steps to this ends, which empirically produced more interpretable output.
First, 
we scale rows of term-document matrix \txmtx 
(corresponding to texter phrasings) 
to unit $\ell_2$ norm before deriving their representation in \latentspace via singular value decomposition. 
Second, we remove the first SVD dimension, i.e., first column of \txrepr, and renormalize each row, before proceeding.

\begin{figure*}
\centering
\includegraphics[width=.85\textwidth]{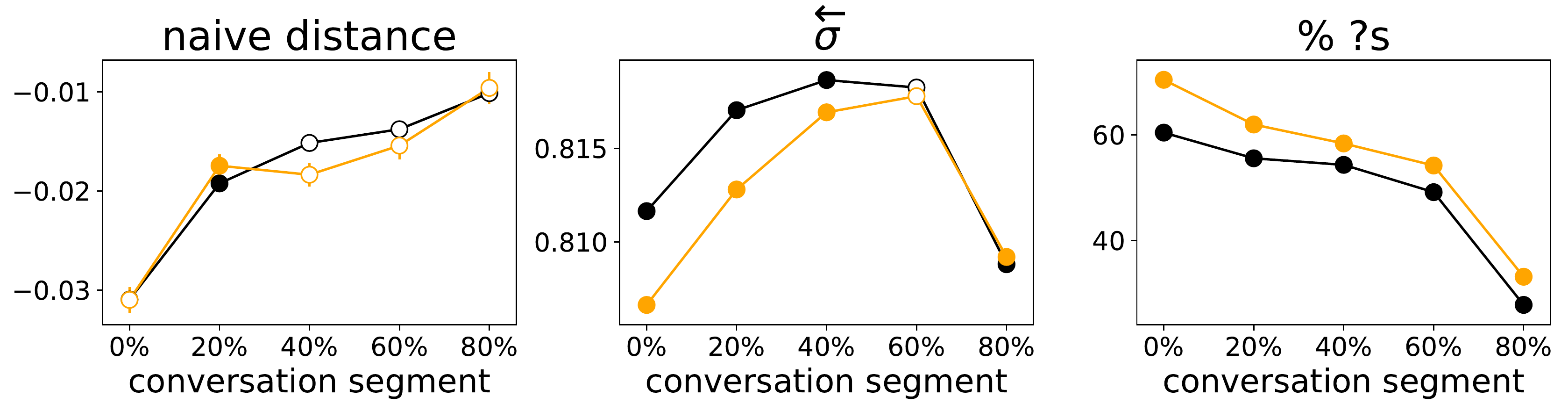}
\caption{
Mean naive distance, backwards-range (\leftattn{}), and \% of utterances with questions, per segment for risk-assessed (orange) and non-risk-assessed (black) conversations; 
solid circles indicate statistically significant differences (Wilcoxon $p < 0.01$, comparing conversation types within counselor).
}
\label{fig:alt_measure_stages}
\end{figure*}

\subsection{Further Details About Application to Counseling Data}
Here, we provide further details on how we applied our methodology to the dataset of counseling conversations in order to measure the orientation of counselor utterances, as described in Section~\ref{sec:validation}. 
In particular, we list empirical choices made in extracting and then processing the training set of 351,935 texter and counselor messages used to measure phrasing orientations.

We randomly sampled 20\% of \textit{counselors} in the data; all conversations by these counselors were omitted in subsequent analyses. 
We merged consecutive messages from the same interlocutor. 
To mitigate potential noise in characterizing phrasings, we considered only 
counselor and texter message pairs in which each message has between 15 and 45 words.
We extracted all messages from the  conversations which met these constraints.

We represent counselor phrasings as dependency-parse arcs and texter messages as unigrams, reflecting the comparatively structured language of the counselors versus the texters (counselors are instructed to speak in grammatically well-formed sentences). 
We consider the 5000 most frequent phrasings for each role, and discard sentences without any such phrasings. 
Finally, we used 25 SVD dimensions to induce \latentspace.

\subsection{Full Listing of Counselor Action Labels}

Table \ref{tab:validation} lists each counseling action derived from the training material and used during the validation procedure (Section \ref{sec:validation_check}) to label sentences.

\begin{table}[]
\small
\begin{tabular}{|l|}
\hline
\begin{tabular}[c]{@{}l@{}}\textbf{reflection} (113)\\ re-wording to show understanding and validate feelings\\ \textit{It can be overwhelming to go through that on your own.}\end{tabular}    \\ \hline
\begin{tabular}[c]{@{}l@{}}\textbf{affirmation} (60)\\ pointing out the texter's positive qualities and actions\\ \textit{You showed a lot of strength in reaching out to us.}\end{tabular}     \\ \hline
\begin{tabular}[c]{@{}l@{}}\textbf{exploration} (44)\\ prompting texters to expand on their situation\\ \textit{Is this the first real fight you've had with your boyfriend?}\end{tabular}      \\ \hline
\begin{tabular}[c]{@{}l@{}}\textbf{problem solving} (110)\\ identifying the texter's goals and potential coping skills\\ \textit{What do you usually do to help you feel calmer?}\end{tabular}  \\ \hline
\begin{tabular}[c]{@{}l@{}}\textbf{closing} (43)\\ reviewing the conversation and transitioning to a close\\ \textit{I think you have a good plan to get some rest tonight.}\end{tabular} \\ \hline
\begin{tabular}[c]{@{}l@{}}\textbf{risk assessment} (19)\\ assessing suicidal ideation or risk of self-harm\\ \textit{Do you have access to the pills right now?}\end{tabular}                  \\ \hline
\end{tabular}
\caption{Counseling strategies and representative examples derived from the training material. 
The number of sentences (out of 400) assigned to each label is shown in parentheses (11 were not labeled as any action).}
\label{tab:validation}
\end{table}

\subsection{Orientation and Lexical Properties}

Here, we supplement our discussion of simple lexical properties that could be used to characterize messages (Section \ref{sec:validation_alt}), 
discussing how orientation reflects these properties and showing that orientation is not subsumed by them.

\xhdr{Backwards-range}
As seen in their weak backwards-range (high, i.e., spread-out \leftattn{}), 
affirmations that highlight the texter's strengths can follow a variety of situations. 
However, the replies they prompt are yet more diffuse, emphasizing the need to compare both directions.

\xhdr{Question-asking} 
We see that questions---which nominally prompt the texter for a response---are more forwards-oriented than non-questions; 
61\% of sentences with `?' have $\orientation{} > 0$ compared to 21\% of sentences without. 
However, these numbers also show that explicitly-marked questions are inexact proxies of forwards-oriented sentences---as in Table \ref{tab:examples},  
questions can address a past remark by prompting clarifications,
while counselors can use non-questions to suggest an intent to advance stages (e.g., to transition to problem-solving).

\subsection{Relating Alternate Measures to Conversation Progress}

Figure \ref{fig:alt_measure_stages} shows averages per conversation segment (i.e., 20\% of a conversation) for each alternative measure considered in Section \ref{sec:validation_alt}. 
Comparing to the average \maxorientation{} and \minorientation{} shown in Figure \ref{fig:val_fig}, 
we see that these measures track with the conversation's progress differently, and none of them distinguish the initial dynamics of risk-assessed conversations as dramatically as reflected in \maxorientation{}, e.g., simple counts of questions do not distinguish between questions geared towards risk-assessment versus more open-ended problem exploration.

\subsection{Further Details About Data Used in Analyses}

Here, we provide further details about the subset of data we used to analyze counselors' orientation behavior (Section \ref{sec:analysis}). 
In particular, our aim was to characterize behavior in typical conversations rather than exceptional cases or those that reflected earlier versions of the training curriculum. 
As such, we only considered the 234,433 conversations which had least five counselor messages, were not risk-assessed or disconnected before completion, and were taken by counselors who joined the platform after January 2017.

\subsection{Counselor-Level Analysis}

Here, we provide further details about our procedure for analyzing counselor-level correspondences between orientation and effectiveness indicators, as alluded to in Section \ref{sec:analysis}.

Recall that our conversation-level findings may be confounded by texter idiosyncracies:
for instance, texters with particularly difficult situations might affect a counselor's behaviour,
but may also be more likely to give bad ratings, independent of how the counselor behaves.
Alternatively, an overly long conversation could arise because the counselor is less forwards-oriented, or because the texter is reluctant to make progress from the outset, making it hard for the counselor to attempt to prompt them forwards. 

To separate a counselor's decisions from these situational effects, we take a counselor-level perspective. 
While counselors cannot selectively talk with different types of texters, they can exhibit cross-conversational inclinations for particular behaviors. 
We therefore relate these cross-conversational \textit{tendencies} in orienting a conversation to a counselor's long-term propensity for receiving helpful ratings, or having long or short  conversations. 
We proceed to describe our methodology for relating counselor tendencies to perceived helpfulness; an analogous procedure could be applied to conversation length as well. 

We characterize a counselor's orienting behavior as the average \maxorientation{} and \minorientation{} over the conversations they take; 
we likewise take the proportion of their (rated) conversations which were perceived as helpful.  
We restrict our counselor level analyses to the 20th to 120th conversations of the 1495 counselors who have taken at least 120 conversations (ignoring their initial conversations when they are still acclimatizing to the platform).

\begin{table*}[]
\small
\begin{tabular}{lll}
\hline
Orientation                                                                   & Example phrasings                                                                                                                        & Example sentences                                                                                                                                                                                                                                                                                                                                                                                                                                                                                \\ \hline
\begin{tabular}[c]{@{}l@{}@{}@{}}Less forwards-\\ oriented\\ (bottom 25\%)\end{tabular} & \begin{tabular}[c]{@{}l@{}}i understand, have been, \\ part of, so you,\\ sentence, talking about \\ might, particular \\ but the, give to \end{tabular}                           & \begin{tabular}[c]{@{}l@{}@{}@{}@{}}As I understand the facts {[}...{]} he had tried to kill the husband, \\ \qquad shooting him twice in the head? (Scalia)\\  You started out by talking about what the first jury knew, \\ \qquad  but {[}...{]} we aren't reviewing that determination. (Roberts)\\ I guess the problem is the list of absurdities that they point to, \\ \qquad  not the least of which is a dry dock. (Sotomayor)\\So you hedged, because it's very hard to find the right sentence. (Breyer)\\\end{tabular} \\ \hline
\begin{tabular}[c]{@{}l@{}@{}@{}}More forwards-\\ oriented\\ (top 25\%)\end{tabular}     & \begin{tabular}[c]{@{}l@{}}hypothetical, would have,\\ agree, difference {[}between{]},\\ {[}do{]} you think, your position\\ your argument, a question \\ apply, was there \end{tabular} & \begin{tabular}[c]{@{}l@{}@{}@{}@{}}Suppose under this hypothetical {[}...{]} the judge doesn't say \\ \qquad aggravated murder when he submits it to the jury. (Kennedy)\\ I just want to know your position on the second, the cart before the \\ \qquad  horse point. (Souter)\\ Do you also agree {[}...{]} that if not properly administered there is some 
\\ \qquad risk of excruciating pain? (Stevens)\\ What's the difference between pigment and color {[}...{]} ? (Ginsburg)\end{tabular}                                     \\ \hline
\end{tabular}
\caption{Example phrasings and sentences from utterances of Supreme Court justices, identified in parentheses, which are less or more forwards-oriented (bottom and top 25\% of \orientation{}).}
\label{tab:examples_scotus}
\end{table*}

To cleanly disentangle counselor tendency and conversational circumstance, we \textit{split} each counselor's conversations into two interleaved subsets (i.e., first, third, fifth $\ldots$ versus second, fourth $\ldots$ conversations), 
measuring orientation on one subset and computing a counselor's propensity for helpful ratings on the other. 
Here, we draw an analogy to the machine learning paradigm of taking a train-test split: ``training'' counselor tendencies on one subset and ``testing'' their relation to rating on the other subset. 
In general, the directions of the effects we observe hold with stronger effects if we do not take this split.

Echoing conversation-level effects, counselors that tend to be less forwards-oriented and more backwards-oriented 
(those in the bottom thirds of \maxorientation{} and \minorientation{} respectively) 
are more likely to be perceived as helpful; 
this contrast is stronger in terms of \minorientation{} (Cohen's $d=0.30$, $p < 0.001$) than \maxorientation{} ($d=0.13$, $p < 0.05$),
suggesting that a counselor's tendency for advancing weighs less on their perceived helpfulness than their tendency for addressing.
Also in line with the conversation-level findings, counselors with smaller \maxorientation{} tend to have longer conversations ($d=0.54, p < 0.001$),
as do counselors with larger \minorientation{} ($d = 0.17$)---here, a counselor's tendency for advancing is more related to their propensity for shorter conversations than their tendency for addressing.

We note that counselors on the platform cannot selectively take conversations with certain texters; 
rather, the platform automatically assigns incoming texters to a counselor. 
As such, the counselor-level effects we observe cannot be explained by counselor self-selection for particular situations.

\subsection{Orientation in Multi-Sentence Utterances}

Our motivation in characterizing utterances using the minimum and maximum sentence orientation was to reflect potential heterogeneities in utterances which could be both backwards- and forwards-oriented 
(consider a message where $c_2$ and $c_1$ from Figure \ref{fig:fig1} are concatenated). 
Examining the 64\% of counselor messages with multiple sentences, we see that 52\% of these messages have $\minorientation{} < 0$ and $\maxorientation{}>0$.
Our method, which is able to account for this heterogeneity, thus points to one potential strategy for counselors to bridge between both objectives. 

\subsection{Application to Supreme Court Oral Arguments}

Here, we include an exploratory study of how our approach could be adapted to analyze domains beyond crisis counseling conversations, as alluded to in Section \ref{sec:discussion}. We apply the method to measure the orientation of utterances by Supreme Court justices during oral arguments, 
when they engage in exchanges with lawyers 
(so justices and lawyers play the roles of counselor and texter, respectively, in our method). 
We used transcripts of 668 cases, taken from the Oyez project (\url{https://www.oyez.org/}), averaging 120 justice utterances per case.\footnote{The data used can be found at \url{http://analysmith.com/research/scotus/data}.} 

Oral arguments contain more linguistic and topical heterogeneiety than counseling conversations, 
since they cover a wide variety of cases, and because the language used by each justice is more differentiated. 
In addition, the dataset is much smaller. 
As such, this represents a more challenging setting than the counseling context, 
requiring changes to the precise procedure used to measure orientation, and pointing to the need for further  technical improvements, discussed in Section \ref{sec:discussion}. 

Nonetheless, our present methodology is able to produce interpretable output. 
Table \ref{tab:examples_scotus} shows representative phrasings and (paraphrased) sentences with different orientations. 
In contrast to the counseling domain, 70\% of phrasings and 93\% of sentences have $\orientation{}  > 0$, 
perhaps reflecting the particular power dynamic in the Supreme Court, 
where justices are tasked with scrutinizing the arguments made by lawyers. 
We find that highly forwards-oriented phrasings and utterances tend to reflect justices pressing on the lawyers to address a point 
(e.g., do you \egword{agree}, what's the \egword{difference between}); 
the least forwards-oriented phrases involve the justice rehashing and reframing (not always in complimentary terms) a lawyer's prior utterances 
(e.g., \egword{so you} [...], [as] \egword{i understand}). 

We used a training set of 15,862 justice and lawyer messages, where each utterance had between 10 and 100 words. Both lawyer and justice utterances were represented as dependency-parse arcs. 
Empirically, we found that the methodology was sensitive to idiosyncracies of particular cases and justices. 
To minimize this effect, we restricted the size of the justice's vocabulary by only considering the 398 justice phrasings which occurred in at least 200 utterances.

\end{document}